\documentclass[a4paper, 11pt]{article}
\pdfoutput=1

\newcommand{\ignore}[1]{}

\usepackage{color,xcolor}
\usepackage{dsfont}
\usepackage{url}
\usepackage{amsmath, amsthm, amssymb}
\usepackage{graphicx,epsfig}
\usepackage{fullpage}

\begin{document}

\title{Predicting the Energy Output of Wind Farms Based on Weather Data: Important Variables and their Correlation}
%\numberofauthors{4} %  in this sample file, there are a *total*
\author{Katya Vladislavleva$^1$, Tobias Friedrich$^2$, Frank Neumann$^3$, Markus Wagner$^3$\\
$^1$ Evolved Analytics Europe BVBA, Veldstraat, 37, 2110, Wijnegem, Belgium\\
$^2$ Max-Planck-Institut f\"ur Informatik, Campus E1.4, 66123 Saarbr\"ucken, Germany\\
$^3$ School of Computer Science, University of Adelaide, Adelaide, SA 5005, Australia}
\date{}

\maketitle
\begin{abstract}
Wind energy plays an increasing role in the supply of energy world-wide. The energy output of a wind farm is highly dependent on the weather condition present at the wind farm. If the output can be predicted more accurately, energy suppliers can coordinate the collaborative production of different energy sources more efficiently to avoid costly overproductions.

With this paper, we take a computer science perspective on energy prediction based on weather data and analyze the important parameters as well as their correlation on the energy output. To deal with the interaction of the different parameters we use symbolic regression based on the genetic programming tool DataModeler. 

Our studies are carried out on publicly available weather and energy data for a wind farm in Australia. We reveal the correlation of the different variables for the energy output. The model obtained for energy prediction gives a very reliable prediction of the energy output for newly given weather data.
\end{abstract}

\section{Introduction}

Renewable energy such as wind and solar energy plays an increasing role in the supply of energy world-wide. This trend will continue because the global energy demand is increasing and the use of nuclear power and traditional sources of energy such as coal and oil is either considered as non-safe or leads to a large amount of CO$_2$ emission.

Wind energy is a key-player in the field of renewable energy. The capacity of wind energy production was increased drastically during the last years. In Europe for example, the capacity of wind energy production has doubled since 2005. 
However, the production of wind energy is hard to predict as it relies on the rather unstable weather conditions present at the wind farm. In particular, the wind speed is crucial for energy production based on wind and the wind speed may vary drastically during different periods of time. Energy suppliers are interested in accurate predictions, as they can avoid overproductions by coordinating the collaborative production of traditional power plants and weather dependent energy sources. 

Our aim is to map weather data to energy production. We want to show that even data that is publicly available for weather stations close to wind farms can be used to give a good prediction of the energy output. Furthermore, we examine the impact of different weather conditions on the energy output of wind farms. We are, in particular, interested in the correlation of different components that characterize the weather conditions such as wind speed, pressure, and temperature

A good overview on the different methods that were recently applied in forecasting of wind power generation can be found in \cite{methodsforcasting2012}. Statistical approaches use historical data to predict the wind speed on an hourly basis or to predict energy output directly. On the other hand, short term prediction is often done based on meteorological data and learning approaches are applied. Kusiak, Zheng, and Song \cite{KusiakPrediction2009} have shown how wind speed data may be used to predict the power output of a wind farm based on times series prediction modeling.
Neural networks are a very popular learning approach for wind power forecasting based on given time series. They provide an implicit model of the function that maps the given weather data to an energy output. 

Jursa and Rohrig~\cite{shorttermforcasting2008} have used particle swarm optimization and differential evolution to minimize the prediction error of neural networks for short-term windpower forecasting. 
Kramer and Gieseke~\cite{KraGieIForcast2011} used support vector regression for short term energy forecast and kernel methods and neural networks to analyze wind energy time series~\cite{KraGieICNC2011}. These studies are all based on wind data and do not take other weather conditions into account. 
Furthermore, neural networks have the disadvantage that they give an implicit model of the function predicting the output, and these models are rarely accessible to a human expert. Usually, one is also interested in the function itself and the impact of the different variables that determine the output.
We want to study the impact of different variables on the energy output of the wind farm. Surely, the wind speed available at the wind farm is a crucial parameter. Other parameters that influence the energy output are for example air pressure, temperature and humidity. Our goal is to study the impact and correlation of these parameters with respect to the energy output. 

Genetic programming (GP) (see \cite{poli08:fieldguide} for a detailed presentation) is a type of evolutionary algorithm that can be used to search for functions that map input data to output data. 
It has been widely used in the field of symbolic regression and the goal of this paper is to show how it can be used for the important real-world problem of predicting energy outputs of wind farms from whether data. The advantage of this method is that it comes up with an \emph{explicit} expression mapping weather data to energy output. This expression can be further analyzed to study the impact of the different variables that determine the output. To compute such an expression, we use the tool DataModeler~\cite{dataModeler:2010:manual} which is the state of the art tool for doing symbolic regression based on genetic programming.
We will use DataModeler also to carry out a sensitivity analysis which studies the correlation between the different variables and their impact on the accuracy of the prediction. 

We proceed as follows. In Section~\ref{sec:gp}, we give a basic introduction into the field of genetic programming and symbolic regression and describe the DataModeler. Section~\ref{sec:approach}, describes our approach of predicting energy output based on weather data and in Section~\ref{sec:results} we report on our experimental results. Finally, we finish with some concluding remarks and topics for future research.

\section{Genetic Programming and DataModeler}
\label{sec:gp}

Genetic programming~\cite{koza:gp2} is a type of evolutionary algorithm that is used in the field of machine learning. Motivated by the evolution process observed in nature computer programs are evolved to solve a given task. Such programs are usually encoded as syntax expression trees. Starting with a given set of trees called the population, new trees called the offspring population are created by applying variation operators such as crossover and mutation. Finally, a new parent population is selected out of the previous parent and the offspring based on how good these trees perform for the given task. 

Genetic programming has its main success stories in the field of symbolic regression. Given a set of input output vectors, the task is to find a function that maps the input to the output as best as possible, while avoiding overfitting. The resulting function is later often used to predict the output for a newly given input.
Syntax trees represent functions in this case and the functions are changed by crossover and mutation to produce new functions. The quality of a syntax trees is determined by how good it maps the given set of inputs to their corresponding outputs.

The task in symbolic regression can be stated as follows. Given a set of data vectors $(x_{1i}, x_{2i}, \ldots, x_{ki}, y_i) \in \mathds{R}^{k+1}$, $1 \leq i \leq n$, find a function $f \colon \mathds{R}^k \rightarrow \mathds{R}$ such that the approximation error, e.g. the root mean square error 
\[
\sqrt{\frac{\sum_{i=1}^n (y_i - f(x_i))^2}{n}}
\]
with $x_i = (x_{1i}, x_{2i}, \ldots, x_{ki})$
is minimized.

We want to use a tool called DataModeler for our investigations. It is based on genetic programming and designed for solving symbolic regression problems.
\subsection{DataModeler}
\label{sec:dm}

Evolved Analytics' DataModeler is a complete data analysis and feature selection environment running under Wolfram Mathematica 8. It offers a platforms for data exploration, data-driven model building, model analysis and management, response exploration and variable sensitivity analysis, model-based outlier detection, data balancing and weighting. 

Data-driven modeling in DataModeler happens by symbolic regression via genetic programming. 
The SymbolicRegression function offers several evolutionary strategies which differ in the applied selection schemes, elitism, reproduction strategies, and fitness evaluation strategies. An advanced user can take full control over symbolic regression and introduce new function primitives, new fitness functions, selection and propagation schemes, etc. by specifying appropriate options in the function call. We, however, used the default settings, and default evolution strategy, called in DataModeler ClassicGP.\footnote{All models reported in this paper were generated using two calls of SymbolicRegression with only the following arguments: input matrix, response vector, execution time, number of independent evolutions, an option to archive models with a certain prefix-name, and a template specification.}

In the symbolic regression performed here a population of individuals (syntax trees) is evolving over a variable number of generations at the Pareto front in the three dimensional objective space of model complexity, model error, and model age~\cite{Kotanchek:2006:GPTP,Schmidt:2010:GPTP}. 

Model error in the default setting ranges between $0$ and $1$ with the best value of $0$. It is computed as $1-R^{2}$, where $R$ is a scaled correlation coefficient. The correlation coefficient of the predicted output is scaled to have the same mean and standard deviation as observed output. 

The model complexity is the expressional complexity of models, and it is computed as the total sum of nodes in all subtrees of the given GP tree. The model age is computed as the number of generations that the model survived in the population. The age of a child individual is computed by incrementing the age of the parent contributing to the root node of the child. We use the age as a secondary optimization objective, as it is used only internally for evolution. At the end of symbolic regression runs results are displayed in the two-objective space of user-selected objectives, in our case these objectives are model expressional complexity and $1-R^{2}$.

The default population size is $300$. The default elite set size is $50$ individuals from the 'old' population closest to the 3-dimensional Pareto front in the objective space. These individuals are copied to the 'new' population of 300 individuals, after which the size of the new population is decreased down to the necessary 300 This is done by selecting models from Pareto layers until the specified amount is found. 

The Selection of individuals for propagation happens by means of Pareto tournaments. By default, $30$ models are randomly sampled from the current population, and Pareto optimal individuals from this sample are determined as winners to undergo variation until a necessary number of new individuals is created. 

\begin{figure}
\centering
\epsfig{file=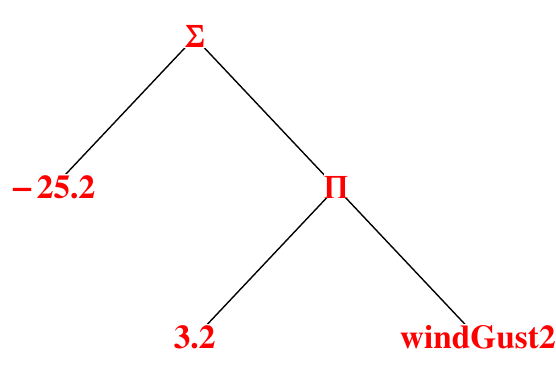, width=2in}
\caption{Model tree plot of the individual from Table~\ref{gpModel}. Model complexity is the sum of nodes in all subtrees of the given tree  ($11$). Model error computed as $1-R^{2}=0.30$.}
\label{gpTree}
\end{figure}

Models are coded as parse-trees using the GPmodel structure, which contains placeholders for information about model quality, data variables and ranges used to develop the model, and some settings of symbolic regression. For example, the internal GPmodel representation of the first Pareto front model from a set of models from Figure~\ref{allModels1} with an expression $-25.2334+3.21666 \text{windGust}_{2}$ is presented in Table~\ref{gpModel}. Note, that the first vector inside the GPmodel structure represents model quality. Model complexity is $11$, model error is $0.300409$. The parse tree of the same model is plotted in Figure~\ref{gpTree}.

\begin{table*}
\centering
\caption{Internal regression model representation in DataModeler for the model with an expression $-25.2334+3.21666 \cdot \text{windGust}_{2}$ (see also Figure~\ref{gpTree}):\smallskip
}{\small\hspace*{-1cm}\begin{tabular}{l }
$\text{GPModel}$ $ [\{11,0.300409\},\Sigma [-25.2334,\Pi [3.21666,\text{windGust2}]],$\\
  $\{\text{ModelAge}\to 1,$  $\text{ModelingObjective}\to \left(\left\{\text{ModelComplexity}[\text{$\#$1}],1-\text{AbsoluteCorrelation}[\text{$\#$2},\text{$\#$3}]^2\right\}\&\right),$\\
  $\text{ModelingObjectiveNames}\to \{\text{Complexity},\text{1-}R^2\},$\\
  $\text{DataVariables}\to \{\text{year},\text{month},\text{day},\text{hour},\text{minute},\text{temperature},\text {apparentTemperature},\text{dewPoint},\text{relativeHumidity},$\\
   $\text{wetBulbDepression},\text{windSpeed}, \text {windGust},\text{windSpeed2},\text{windGust2},\text{pressureQNH},\text{rainSince9am}\},$\\
  $\text{DataVariableRange}\to \{\{2010,2011\},\{1,12\},\{1,31\},\{0,23\},\{0,30\},\{4.2,23.4\},\{-14.2,24.\},\{-3.2,19.1\},\{40,100\},$\\
  $\{0.,6.6\},\{0,106\},\{0,130\},\{0,57\},\{0,70\},\{987.8,1037.5\},\{0.,50.4\}\},\text{RangeExpansion}\to \text{None},$\\
  $\text{ModelingVariables}\to \{\text{year},\text{month},\text{day},\text{hour},\text{minute},\text{temperature},\text{apparentTemperature},\text{dewPoint},\text{relativeHumidity},$\\
  $\text{wetBulbDepression},\text{windSpeed},\text{windGust},\text{windSpeed2},\text{windGust2},\text{pressureQNH},\text{rainSince9am}\},$\\
   $\text{FunctionPatterns}\to \{\Sigma [\_,\_\_],\Pi [\_,\_\_],\mathbb{D}[\_,\_],\mathbb{S}[\_,\_],\text{$\mathbb{P}$2}[\_],\mathbb{S}\mathbb{Q}[\_],\mathbb{I}\mathbb{V}[\_],\mathbb{M}[\_]\},\text{StoreModelSet}\to \text{True},$\\
   $ \text{ProjectName}\to \text{fullDataAllVars},\text{TemplateTopLevel}\to \{\Sigma [\_,\_\_]\}, \text{TimeConstraint}\to 2000,\text{IndependentEvolutions}\to 10\}]$.   
\end{tabular}}
 \label{gpModel}
\end{table*}

When a specified execution threshold of a run in seconds is reached, the independent evolution run terminates and a vector of model objectives in the final population is re-evaluated to only contain model complexity and model error. The set of models can further be analyzed for variable drivers, most frequent variable combinations, behavior of the response, consistency in prediction, accuracy vs. complexity trade-offs, etc. 

When the goal is the prediction of the output in the unobserved region of the data space, it is essential to use 'model ensemble' rather than individual models for this purpose. Because of built-in niching, complexity control, and independent evolutions used in DataModeler's symbolic regression, the final models are developed to be diverse (with respect to structural complexity, model forms, residuals), but they all are global models, built to predict training response in the entire training region. Due to diversity and high quality, rich sets of final models allow us to select multiple individuals to model ensembles. Prediction of a set of individuals is then computed as a median or a median average of individual predictions of ensemble members, while disagreement in the predictions (standard deviation in this paper) is used to specify the confidence interval of prediction. When models are extrapolated, the confidence of predictions naturally deteriorates and confidence intervals become wider. This allows first, a more robust prediction of the response (since over-fitting is further mitigated by choosing models of different accuracy and complexity into an ensemble), and second, it makes the predictions more trustworthy, since predictions are also supplied with confidence intervals. 

To select ensembles we used a built-in function of DataModeler, that focuses on most typical individuals of the model set as well as on individuals that have least correlated residuals. Because of space constraints we refer the reader to \cite{dataModeler:2010:manual} for further information.

\section{Our Approach}
\label{sec:approach}

The main goal of this paper is to use public data to check feasibility of wind energy prediction by using a industrial-strength off-the-shelf non-linear modeling and feature selection tool. 
In our study, we investigate and predict the energy production of the wind farm \emph{Woolnorth} in Tasmania, Australia based on publicly available data. The energy production data is made publicly available by the Australian Energy Market Operator (AEMO) in real time to assist in maintaining the security of the power system.\footnote{Australian Landscape Guardians: AEMO Non-Scheduled Generation Data: \url{www.landscapeguardians.org.au/data/aemo/} (last visited August 31st, 2011)}  
For the creation of our models and the prediction, we associate the wind farm with the Australian weather station \emph{ID091245}, located at Cape Grim, Tasmania. Its data is available for free for a running observation time window of 72 hours.\footnote{Australian Government, Bureau of Meteorology: weather observations for Cape Grim: 
\url{www.bom.gov.au/products/IDT60801/IDT60801.94954.shtml}
 (last visited August 31st, 2011)}

\subsection{Data}
We collected both the weather and energy production data for the time window September 2010 till July 2011. 
The output of the farm is available with a rate of one measurement every five minutes, and the weather data with a rate of one measurement every 30 minutes.

The wind farm's production capacity is split into two sites, which complicated the generation of models. The site "Studland Bay" has a maximum output of 75 MW, and "Bluff Point" has a maximum output of 65 MW and is located 50km south of the first site. The weather station is located on the first site. %, which made the modelling complex.
For wind coming from west (which is the prevailing wind direction), the difference in location is negligible. But if wind comes from north, there will be an energy and wind increase right away, plus another energy increase 1-2 hours later (the time delay depends on the actual wind speed). Similarly, if wind comes from south, there will be an increase in the energy production (although no wind is indicated by the weather station) and then 1-2 hours later an energy increase accompanied by a measured wind speed increase.

\subsection{Data pre-processing}
To perform data modeling and variable selection on collected data, we had to perform data pre-processing to create a table of weather and energy measurements taken at the same time intervals.  
Energy output of the farm is measured at the rate of 5 minutes, including the time stamps of 0 and 30 minutes of every hour when the weather is measured. Our approach was to correlate weather measurements with the average energy energy output of the farm reported in the $[0, 25]$ and $[30,55]$ minute intervals of every hour. Such averaging makes modeling more difficult, but uses all energy information available. 

Different time scales used in the weather and energy data were automatically converted to one scale using a DateList
 function of Wolfram Mathematica~8, which is the scientific computing environment in which DataModeler operates. 

Because of many missing, erroneous, and duplicate time stamps in the weather data we obtained $11022$ common measurements of weather and averaged energy produced by the farm from October 2010 till June 2011. These samples were used as training data to build regression models. From $18$ variables of the weather data at Cape Grim we excluded two variables prior to modeling: the \emph{Pressure MSL} variable had more than 75\% missing values and the \emph{Wind Direction} variable was non-numeric. 

As test data we used $1408$ common half-hour measurements of weather and averaged energy in July 2011. 

\subsection{Data Analysis and Model Development}

As soon as weather and energy data from different sources were put in an appropriate input-output form, we were able to apply a standard data-driven modeling approach to them. 

A good approach employs iterations between three stages: Data Collection/Reduction, Model Development, and Model Analysis and Variable Selection. In hard problems many iterations are required to identify a subspace of minimal dimensionality where models of appropriate accuracy and complexity trade-offs can be built. 

\begin{figure}
\centering
\epsfig{file=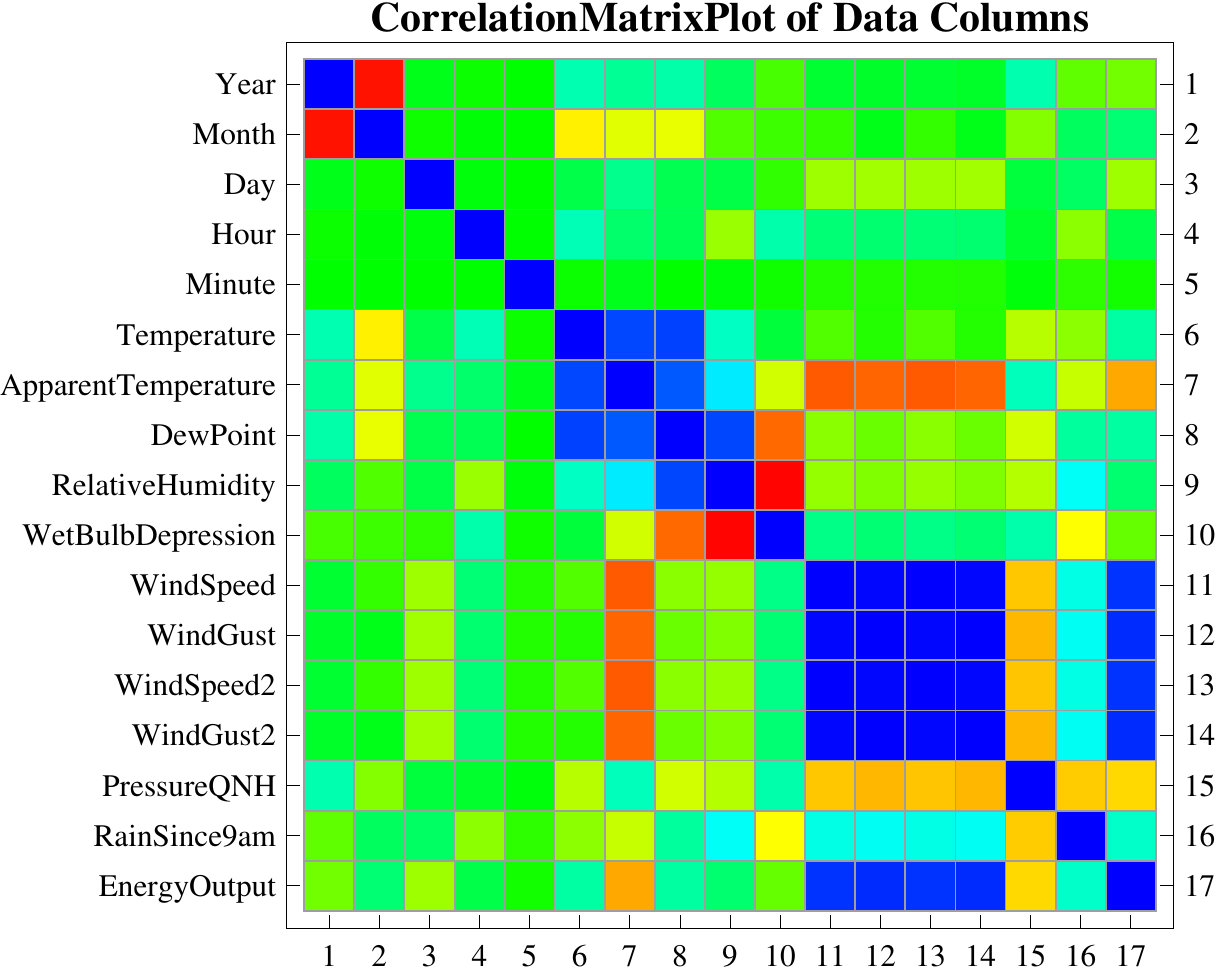, width=3in}
\caption{Data variables are heavily correlated (Blue - positively, Red - negatively).}
\label{correlationMatrixPlot}
\end{figure}

Our problem is challenging for several reasons. First, it is hard to predict the total wind energy output of the farm in half an hour following the moment when weather is measured, especially when the weather station is several kilometers away from the farm). Second, public data does not offer any information about the wind farm except for wind energy output. Third, our training data covers the range of weather conditions observed only between October 2010 and June 2011 while the test data contains data from July implying that our models must have good generalization capabilities as they will be extrapolated to the unseen regions of the data space. And lastly, our challenging goal is to use all $16$ publicly available numeric weather characteristics for energy output prediction, while many of them are heavily correlated (see Table \ref{correlationMatrixPlot}). 

Multi-collinearity in hard high-dimensional problems is a major hurdle for most regression methods. Symbolic regression via GP is one of the very few methods which does not suffer from multicollinearity and is capable of naturally selecting variables from the correlated subset for final regression models. 

Because ensemble-based symbolic regression and robust variable selection methodology are implemented in DataModeler we settled for a standard model development and variable selection procedures using default settings. 

The modeling goals of this study are: 
\begin{enumerate}
\item to identify the minimal subset of driving weather features that are significantly related to the wind energy output of the wind farm, 
\item to let genetic programming express these relationships in the form of explicit input-output regression models, and
\item to select model ensembles for improved generalization capabilities of energy predictions and to analyze the quality of produced model ensembles using an unseen test set.
\end{enumerate}

Our approach is to achieve these goals using two iterations of symbolic regression modeling.  At the first exploratory stage we run symbolic regression on training data to identify driving weather characteristics significantly related to the energy output. At the second modeling stage we reduce the training data to the set of selected inputs and run symbolic regression to obtain models, and model ensembles for predicting energy output.

\section{Experimental Results}
\label{sec:results}

\subsection{Experimental setup}

The setup of symbolic regression used default settings of DataModeler except for the number of independent runs, execution time of each run and the template operator at the root of the GP trees. We executed 10 independent evolutionary runs of  $2000$ seconds in both stages. The root node of all GP trees was fixed to a Plus. The primitives for regression models consisted of an extended set of arithmetic operators: $\{\text{Plus}, $ $\text{Minus}, \text{Subtract}, \text{Divide}, \text{Times}, \text{Sqrt}, \text{Square}, \text{Inverse}\}$. The maximum arity of Plus and Times operators is limited to $5$.  

\begin{figure}
\centering
\epsfig{file=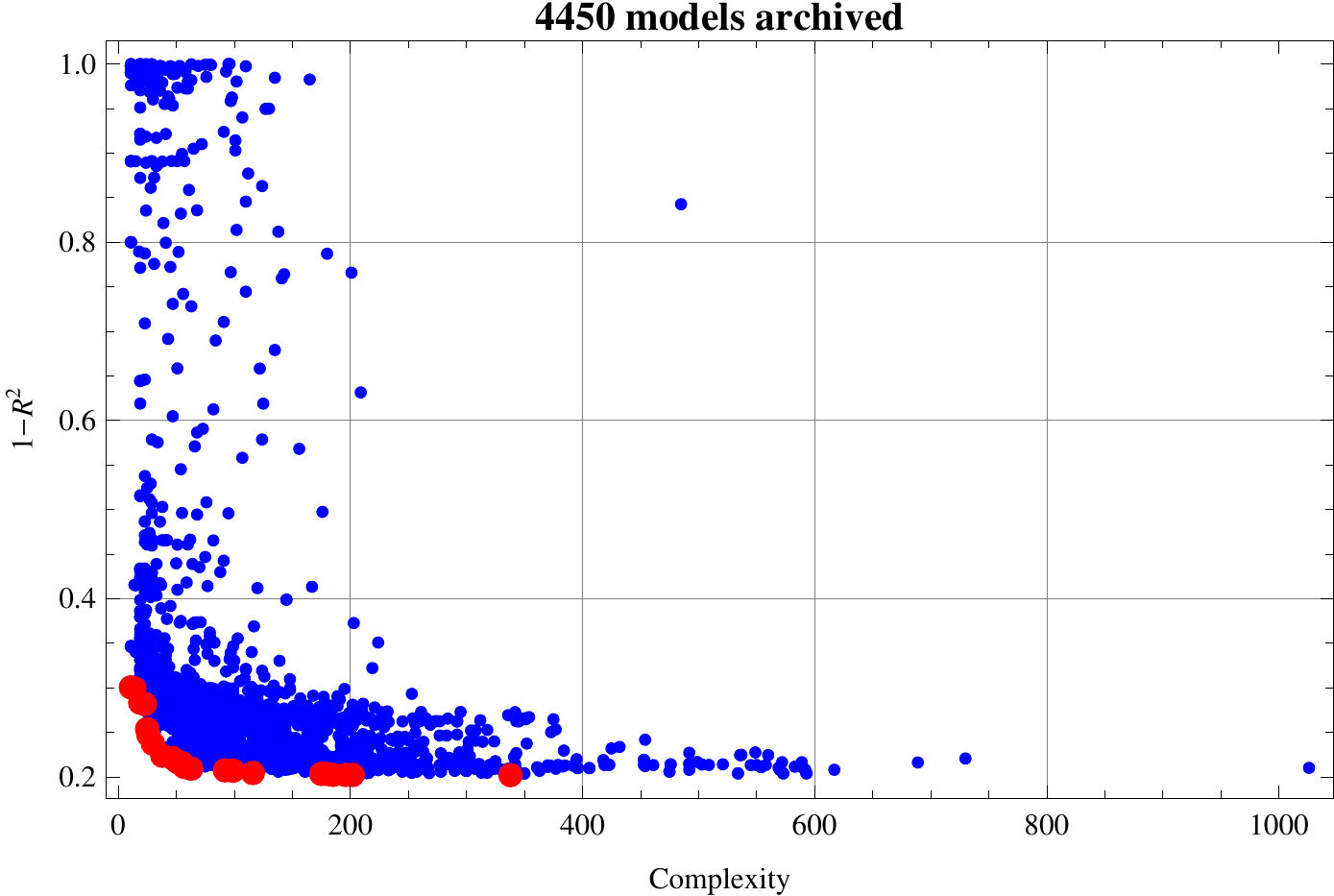, width=3in}
\caption{A super set of models generated in the first stage of experiments with 10 independent evolutions using all inputs. Red dots are Pareto front models, which are non-dominated trade-offs in the space of model complexity and model error.}
\label{allModels1}
\end{figure}

Model trees have terminals labelled as variables or constants (random integers or reals), with a maximum allowed model complexity of $1000$. Population size is $300$, elite set size is $50$. Population individuals are selected for reproduction using Pareto tournaments with the tournament size of $30$. Propagation operators are crossover (at rate $0.9$), subtree mutation (rate $0.05$), and depth preserving subtree mutation (rate $0.05$). At the end of each independent evolution the population and archive individuals are merged together to produce a final set of models. At each stage of experiments the results of all independent evolutions are merged together to produce a super set of solutions (see an example in Figure \ref{allModels1}).

For model analysis we applied additional model selection strategies to these super sets of models. 
We describe the additional model selection strategies, discovered variable drivers, final models, and the quality  of predictions in the next section.

\subsection{Feature selection}

The initial set of experiments targets the feature selection, using all $16$ input variables and all training data from October 2010 till June 2011. In the allowed $2000$ seconds each symbolic regression run completed at most 217 generations. 

\begin{figure}
\centering
\epsfig{file=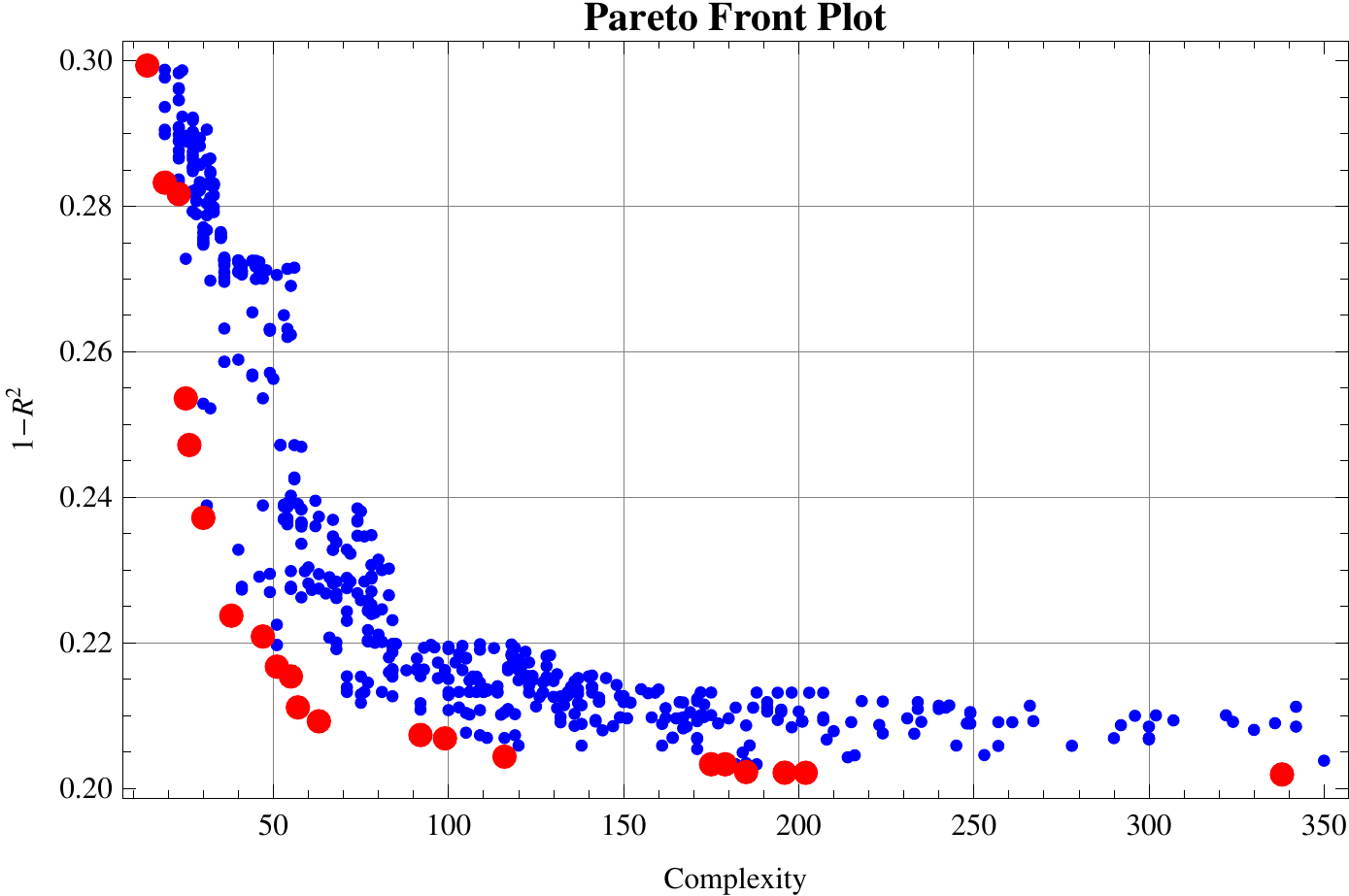, width=3in}
\caption{Selected set $\mathcal{M}_{1}$ of 'best' models in all variables and two modeling objectives.}
\label{selectedModels}
\end{figure}

The 10 independent evolutions generated a super set of $4450$ models. We reduced this set to robust models only, by applying interval arithmetic to remove models with potential for pathologies and unbounded response in the training data range. This generated $2559$ unique robust models, and from those we selected the final set $\mathcal{M}_{1}$. This set contained $587$ individuals with the model error not exceeding $0.30$, and model complexity not exceeding $350$, which lie closest to the Pareto front in Model Complexity versus Model Error objective space. The set $\mathcal{M}_{1}$ is depicted in Figure~\ref{selectedModels} with Pareto front individuals indicated in red. The limit $350$ on Model Complexity preserved the best of the run model (the right-most red dot), but excluded dominated individuals with model complexities up to $600$. 

\begin{figure}
\centering
\epsfig{file=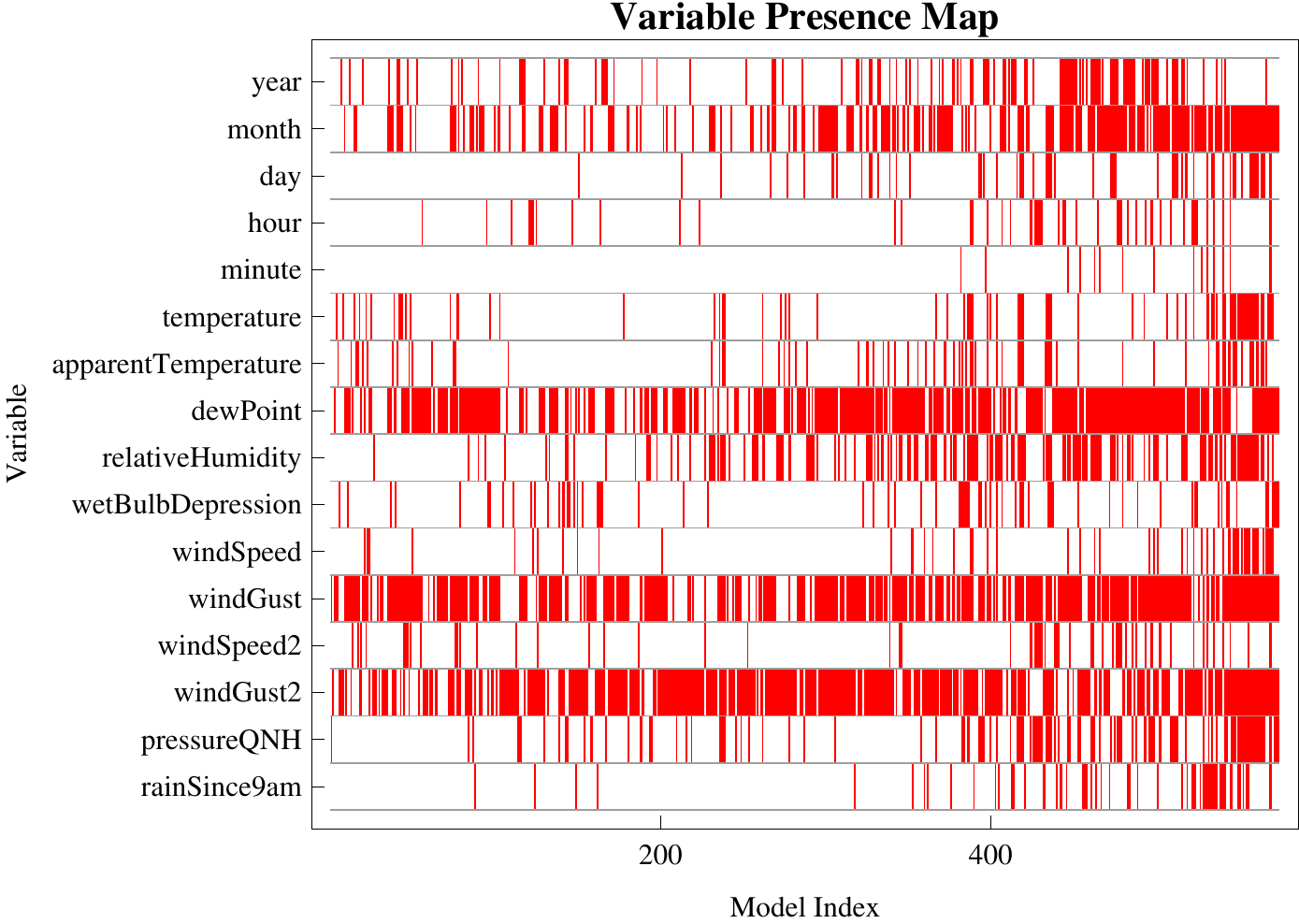, width=3in}
\caption{Presence of input variables in the selected set $\mathcal{M}_{1}$.}
\label{VariablePresenceMap}
\end{figure}

\begin{figure}
\centering
\epsfig{file=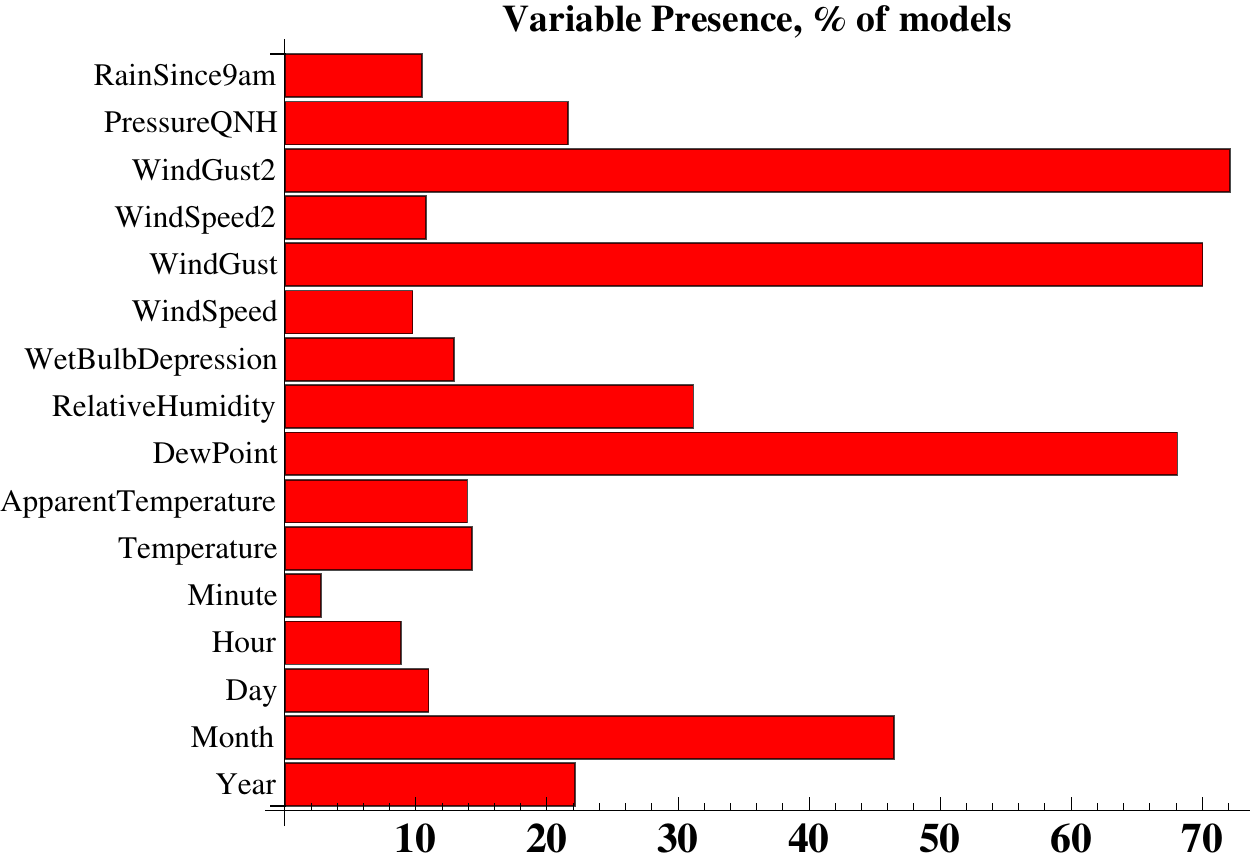, width=3in}
\caption{Presence of input variables in the selected set of models.}
\label{VariablePresence}
\end{figure}

We used the set $\mathcal{M}_{1}$ to perform variable presence and variable contribution analysis to identify the variable drivers significantly related to energy output. The presence of input variables in models from $\mathcal{M}_{1}$ is vizualised in Figure~\ref{VariablePresenceMap} and Figure~\ref{VariablePresence}. We can observe from Figure~\ref{VariablePresence} that the six most frequently used variables are (in the order of decreasing importance) windGust$_{2}$, windGust, dewPoint, month, relativeHumidity, and pressureQNH. While we observe that these variables are most frequently used in a good set of candidate solutions in $\mathcal{M}_{1}$, it is somewhat hard to define a threshold on these presence-based variable importances to select variable drivers. For example it is unclear whether we should select the top three, top four, or top five inputs.

\begin{figure}
\centering
\epsfig{file=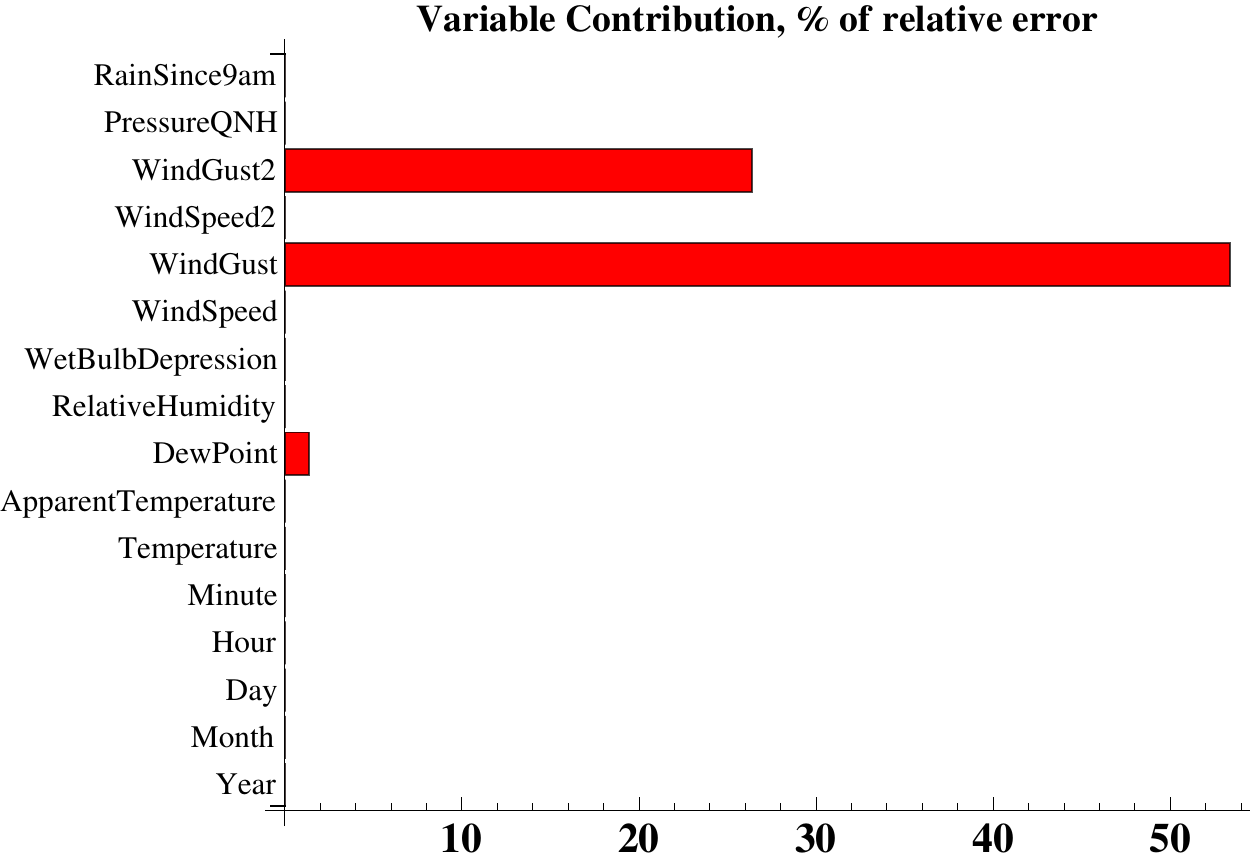, width=3in}
\caption{Individual contributions of input variables in the selected set of models to the relative training error.}
\label{VariableContribution}
\end{figure}

For a crisper feature selection analysis we performed a variable contribution analysis using DataModeler to see how much contribution does each variable have to the relative error of the model where it is present. The median variable contributions computed using the model set $\mathcal{M}_{1}$ are depicted in Figure~\ref{VariableContribution}. The plot clearly demonstrates that the contribution is negligible of other variables besides the top three mentioned above and identified using variable presence analysis. 

Results of the first stage of experiments suggest that the weather inputs windGust2, windGust, and DewPoint are 1) the most frequently present in $\mathcal{M}_{1}$ and 2) have the highest contribution to the relative errors of models in $\mathcal{M}_{1}$ and are sufficient to achieve the accuracy of $\mathcal{M}_{1}$. In other words these inputs are sufficient to predict energy output with accuracy between $70\%$ and $80\%$ $R^{2}$ on the training data. 

The high correlation between windGust and windGust2 variables motivated us to select only one of them for the second round of modeling together with dewPoint to generate prediction models. Symbolic regression does not guarantee that only one particular input variable out of the set of correlated inputs will be present in final models. It might be that either only one out of two is sufficient to predict the response with the same accuracy, or that both are necessary for success. Our choice was to select the windGust2  (as the most frequent variable in the models) together with dewPoint for the second stage of experiments and see whether predictive accuracy of new models in the new two-dimensional design space will not get worse, when compared to the accuracy of $\mathcal{M}_{1}$ models developed in the original space of $16$ dimensions. 

\subsection{Energy output prediction}

\begin{figure}
\centering
\epsfig{file=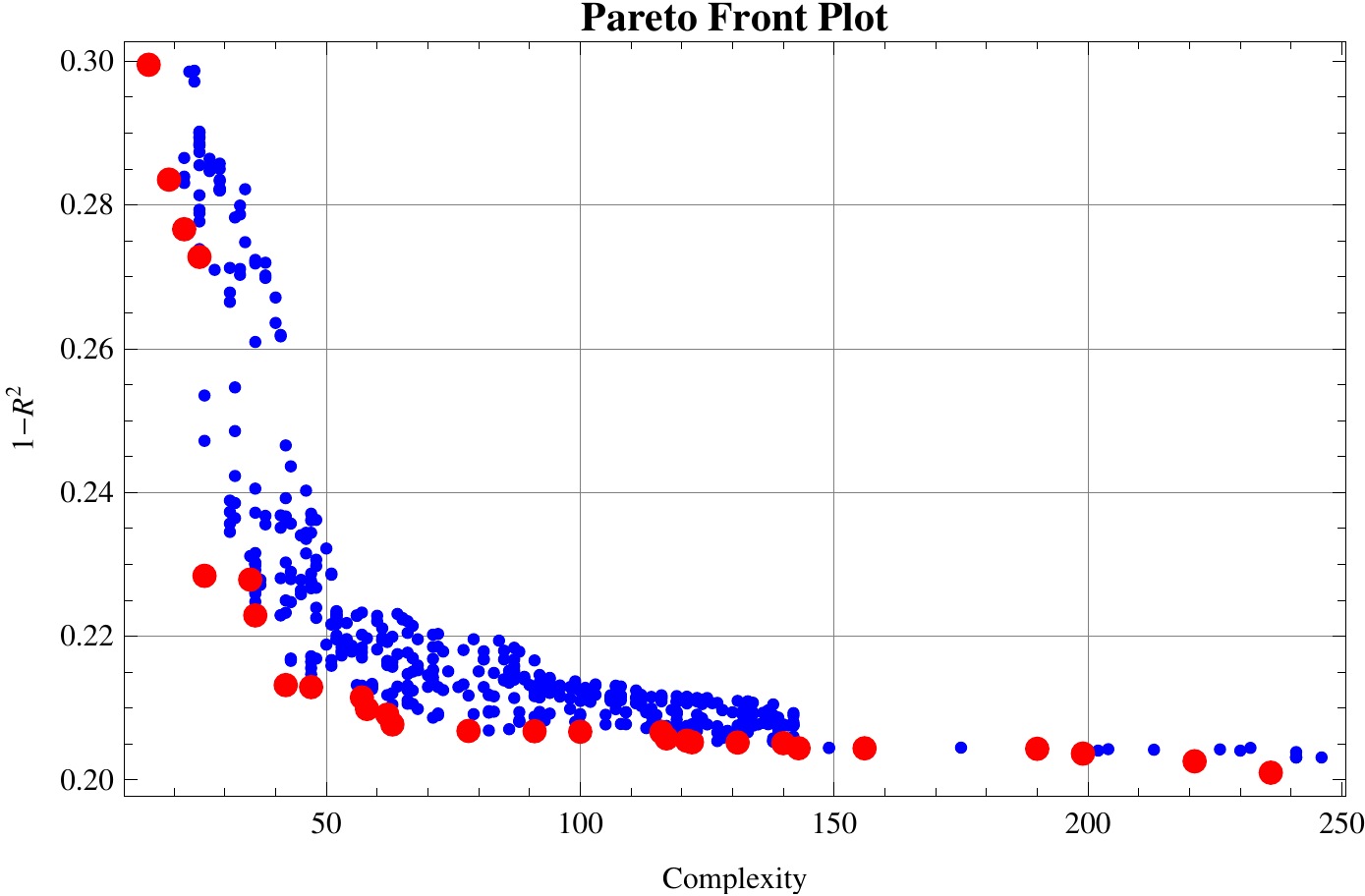, width=3in}
\caption{Selected set $\mathcal{M}_{2}$ of 'best' models in up to two-dimensional input space and two modeling objectives.}
\label{M2}
\end{figure}

\begin{figure}
\centering
\epsfig{file=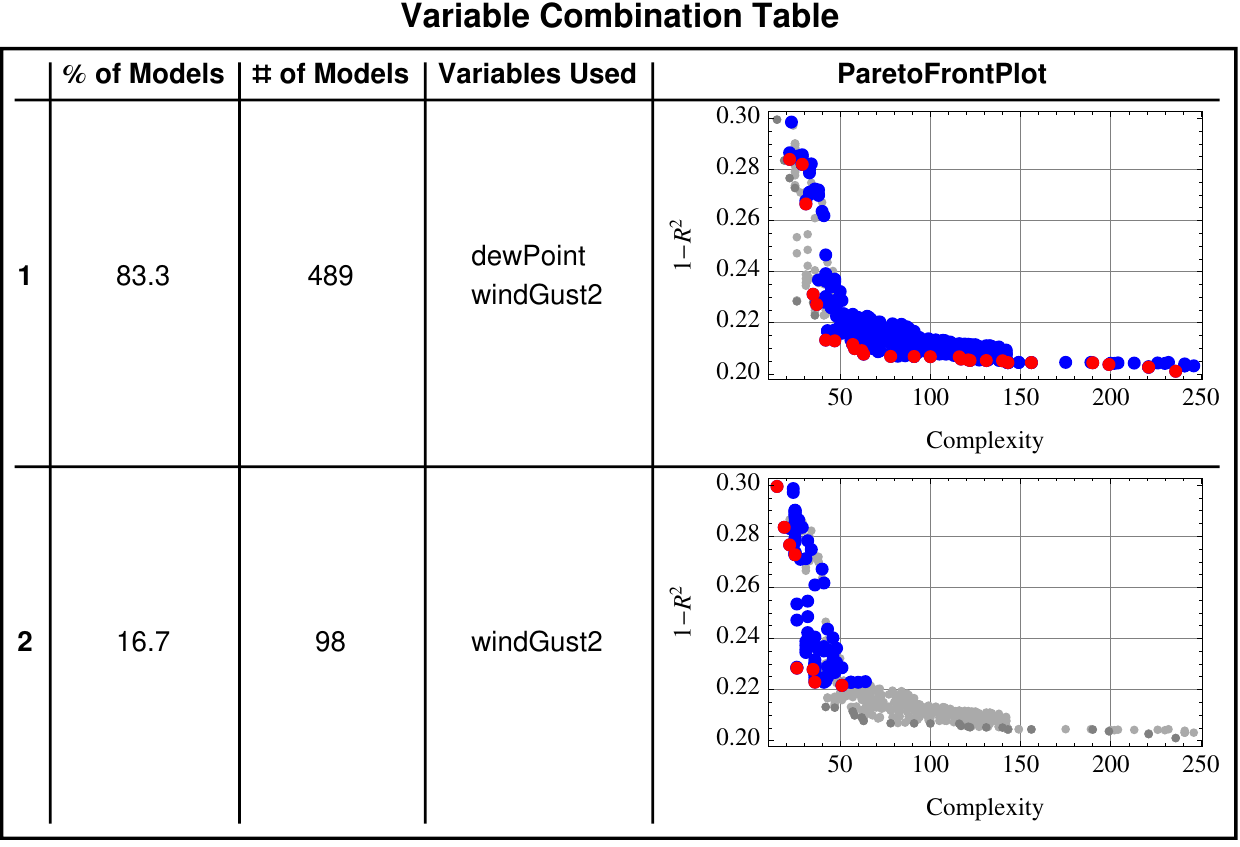, width=3in}
\caption{Visualization of models in $\mathcal{M}_{2}$ niched per driving variable combination. Note, that windGust2 alone is insufficient to predict energy output with the accuracy that is achieved when windGust2 and dewPoint are used. The model error is computed using training data.}
\label{M2VCT}
\end{figure}

The second stage of experiments used only the two input variables windGust2 and dewPoint, with all other symbolic regression settings identical to the first stage experiments. As a result, a new set of one and two-variable models was generated. We again applied a selection procedure to the superset of models by selecting only 25\% of robust models closest to the Pareto front with the training error of at most $1-R^{2}=0.30$ and model complexity of at most $250$. The resulting set of $587$ simplest models, denoted as $\mathcal{M}_{2}$ is depicted in Figures~\ref{M2} and~\ref{M2VCT}. 

Figure~\ref{M2VCT} is obtained using the VariableContributionTable function of DataModeler, and it exposes the trade-offs for input spaces and prediction accuracy for energy prediction. 

We emphasize here that this is the decision and the responsibility of the domain expert to pick an appropriate input space for the energy prediction models. This decision will be guided by the costs and risks associated with different prediction accuracies, and by the time needed to perform measurements of associated design spaces. The responsibility of a good model development tool is to empower experts with robust information about the trade-offs. 

At the last stage of model analysis we used the CreateModelEnsemble function of DataModeler to select an ensemble of regression models from $\mathcal{M}_{2}$ but only allowing models with model complexities not exceeding $150$. As can be seen in Figure~\ref{M2}, an increase of model complexity does not provide a sufficient increase in the training error. Since our goal is to predict energy production on a completely new interval of weather conditions (here: July 2011) we settle for the simplest models to avoid potential over-fitting. 

\begin{table*}
\centering
\caption{Model Ensemble (six models) selected from $\mathcal{M}_{2}$. Constants are rounded to one digit after comma.\smallskip}
{\small\hspace*{-1.5cm}\begin{tabular}{l }
      $-32.1+2.9 \left(\sqrt{\text{windGust}_2}+\text{windGust}_2\right)$ \\
 $112.0 -3.5*10^{-5} \left(-1956.3+\text{dewPoint}^2+\text{windGust}_2{}^2\right){}^2$\\
  $-6.4+1.3*10^{-4} \left(9-\sqrt{\text{windGust}_2}\right){}^2 \text{windGust}_2{}^2 \left(-9.9+\text{dewPoint}+2 \text{windGust}_2\right)$\\
   $-4.5+4.3*10^{-4} \left(-8.9+\sqrt{\text{windGust}_2}\right) \left(-\sqrt{\text{windGust}_2}+0.1 \text{windGust}_2\right) \text{windGust}_2 \left(-12+\text{dewPoint}^2+\text{windGust}_2{}^2\right)$\\
    $-3.1+1.5*10^{-4} \left(-3 \text{ dewPoint }  \text{windGust}_2{}^2+\left(9-\sqrt{\text{windGust}_2}\right)^2 \text{windGust}_2{}^2 \left(-16.3+\text{dewPoint}+2 \text{windGust}_2\right)\right)$\\
     $-11.2+9.4*10^{-7} \left(9-\sqrt{\text{windGust}_2}\right){}^2 \sqrt{\text{windGust}_2} \left(39.4 +4 \text{dewPoint}+7 \text{windGust}_2\right) \left(\frac{1}{9}+\text{dewPoint}+\left(10+2 \text{windGust}_2\right){}^2\right)$\\
\end{tabular}}\vspace{-2mm}
\label{ensemble}
\end{table*}

The selected model ensemble consists of six models presented in Table~\ref{ensemble}.  The values of model complexity, training error, and test error\footnote{Test error is, of course, evaluated post facto, after the models are selected into the model ensemble.} for six models in the ensemble are respectively $(24,0.299, 0.426 )$, $(42,0.247, 0.472)$, $(63,0.209, 0.146)$, $(78,0.207, 0.149)$, $(121,0.205, 0.145)$, $(124,0.211, 0.145)$.

The created model ensemble can now be evaluated on the test data. As mentioned in Section~\ref{sec:dm} ensemble prediction is computed as a median of predictions of individual ensemble members, while ensemble confidence is computed as a standard deviation of individual predictions. We report that the normalized root mean squared error of ensemble prediction on the test data is $\text{RMSE}_\text{Test}=12.6\%.$ 

\begin{figure}
\centering
\epsfig{file=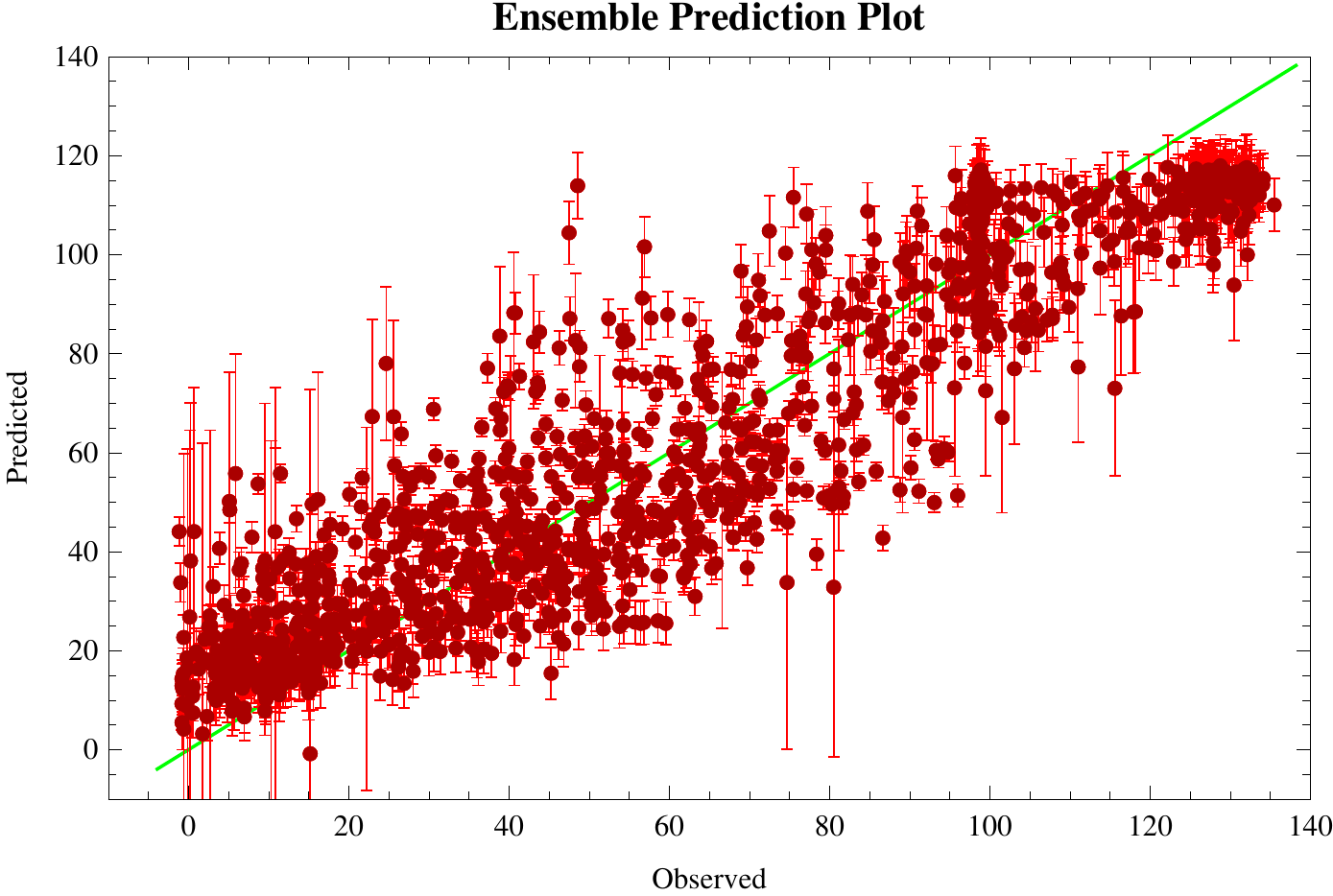, width=3in}
\caption{Ensemble prediction versus observed energy output in July  (Test Data) of the final model ensemble. Whiskers correspond to ensemble disagreement measured as a standard deviation between predictions of individual ensemble members for any given input sample.}
\label{PredictedVsActual}
\end{figure}

\begin{figure*}
\centering
\epsfig{file=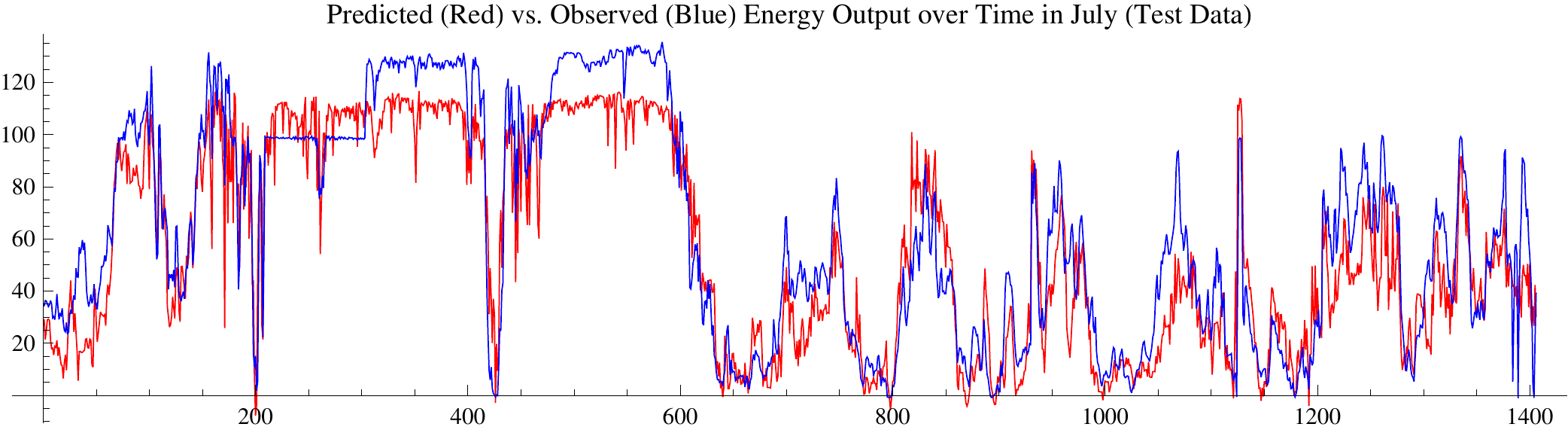, width=6in}
\caption{Ensemble Prediction versus Actual energy output  over time on the Test Data.}
\label{PredictedVsActualTime}
\end{figure*}

Figure~\ref{PredictedVsActual} presents the predicted versus observed energy output in July 2011, with whiskers corresponding to ensemble confidence. Note that the confidence intervals for prediction are very high for many training samples. This is normal and should be expected when prediction is evaluated well beyond the training data range. Figure~\ref{PredictedVsActualTime} presents ensemble prediction versus actual energy production over time in July 2011. 

\section{Conclusions}
In this study we showed that wind energy output can be predicted from publicly available weather data with accuracy at best 80\% $R^{2}$ on the training range and at best $85,5\%$ on the unseen test data.  We identified the smallest space of input variables (windGust2 and dewPoint), where reported accuracy can be achieved, and provided clear trade-offs of prediction accuracy for decreasing the input space to the windGust2 variable. We demonstrated that an off-the-shelf data modeling and variable selection tool can be used with mostly default settings to run the symbolic regression experiments as well as variable importance, variable contribution analysis, ensemble selection and validation.  

We are looking forward to discuss the results with domain experts and check the applicability of produced models in real-life for short term energy production prediction. We are glad that the presented framework is so simple that it can be used literally by everybody for predicting wind energy production on a smaller scale---for individual wind mills on private farms or urban buildings, or small wind farms. 
For future work, we are planning to study further the possibilities for longer-term wind energy forecasting. 

\bibliographystyle{abbrv}
\bibliography{arxiv}  

\end{document}